\documentclass[11pt,table]{article} 

\usepackage{acl}

\usepackage{times}
\usepackage{latexsym}
\usepackage{makecell} 
\usepackage{booktabs}
\usepackage{amssymb}
\usepackage{placeins}

\usepackage[T1]{fontenc}

\usepackage[utf8]{inputenc}

\usepackage{microtype}

\usepackage{inconsolata}
\usepackage{makecell}
\usepackage{graphicx}

\title{KRETA: A Benchmark for Korean Reading and Reasoning in Text-Rich VQA Attuned to Diverse Visual Contexts}

\author{
  Taebaek Hwang\textsuperscript{1}\thanks{Equal contribution.} \quad
  Minseo Kim\textsuperscript{2}\footnotemark[1] \quad
  Gisang Lee\textsuperscript{3} \quad
  Seonuk Kim\textsuperscript{4} \quad
  Hyunjun Eun\textsuperscript{5} \\
  \\
\textsuperscript{1}Waddle \quad
\textsuperscript{2}Seoul National University \quad
\textsuperscript{3}Krafton \quad
\textsuperscript{4}UNIST \quad
\textsuperscript{5}SK Telecom
}

\begin{document}
\maketitle
\begin{abstract}
Understanding and reasoning over text within visual contexts poses a significant challenge for Vision-Language Models (VLMs), given the complexity and diversity of real-world scenarios. To address this challenge, text-rich Visual Question Answering (VQA) datasets and benchmarks have emerged for high-resource languages like English. However, a critical gap persists for low-resource languages such as Korean, where the lack of comprehensive benchmarks hinders robust model evaluation and comparison. To bridge this gap, we introduce \textbf{KRETA}, a benchmark for \textbf{K}orean \textbf{R}eading and r\textbf{E}asoning in \textbf{T}ext-rich VQA \textbf{A}ttuned to diverse visual contexts. KRETA facilitates an in-depth evaluation of both visual text understanding and reasoning capabilities, while also supporting a multifaceted assessment across 15 domains and 26 image types. Additionally, we introduce a semi-automated VQA generation pipeline specifically optimized for text-rich settings, leveraging refined stepwise image decomposition and a rigorous seven-metric evaluation protocol to ensure data quality. While KRETA is tailored for Korean, we hope our adaptable and extensible pipeline will facilitate the development of similar benchmarks in other languages, thereby accelerating multilingual VLM research. The code and dataset for KRETA are available at \href{https://github.com/tabtoyou/KRETA}{\texttt{https://github.com/tabtoyou/KRETA}}.

\vspace{-0.7em}
\end{abstract}

\begin{figure*}
  \centering 
  \includegraphics[width=0.94\textwidth]{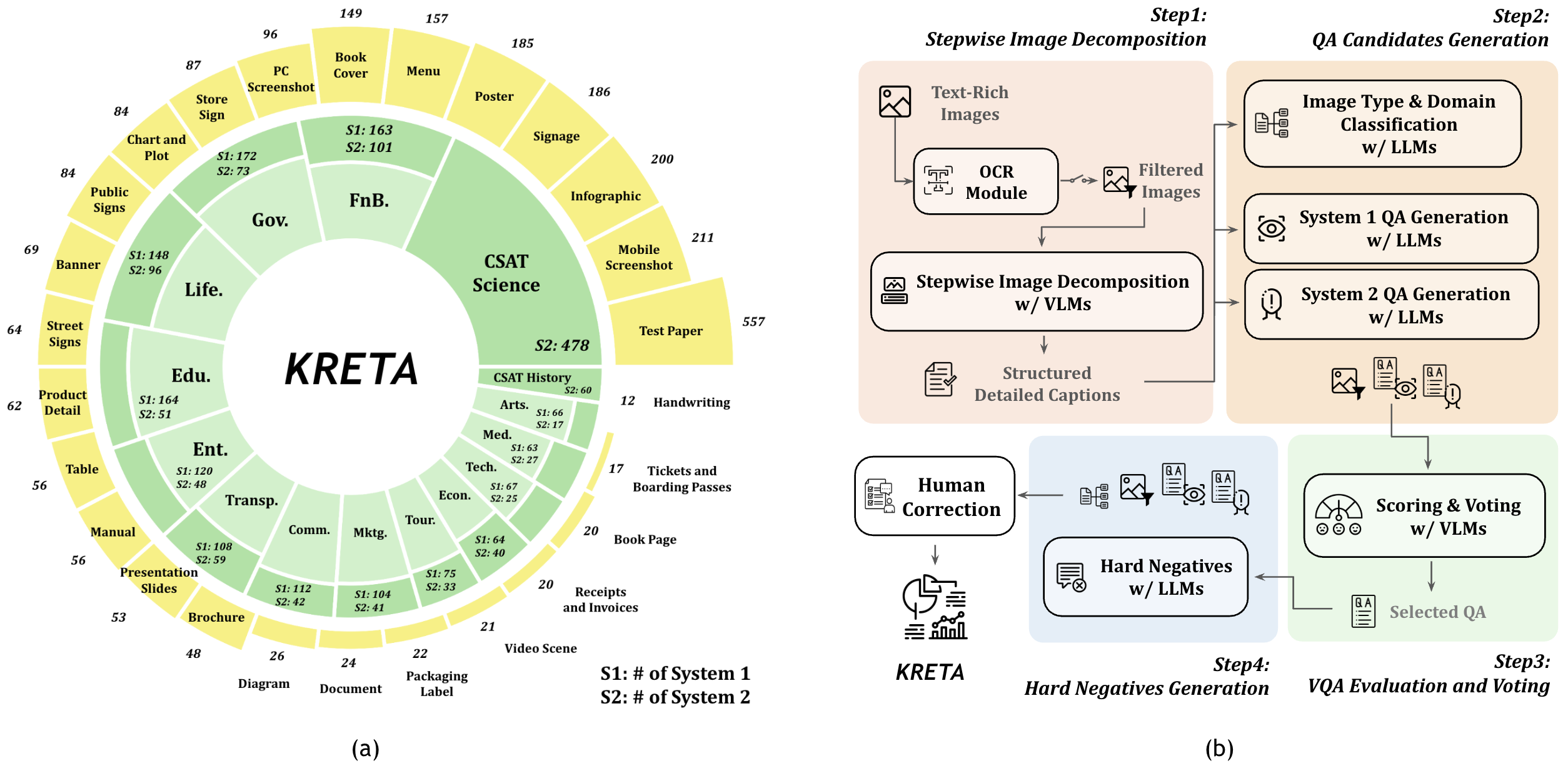}
  \vspace{-0.5em}
  \caption{(a) Distribution of samples across 15 domains (inner ring) and 26 image types (outer ring). Dark green and light green segments in the inner ring represent the number of samples associated with System 2 and System 1, respectively. See Subsection~\ref{ref:domain} for domain abbreviations. (b) The semi-automated VQA generation pipeline.}
  \label{fig:overview-of-kreta}
  \vspace{-1.2em}
\end{figure*}

\section{Introduction}

In real-world scenarios, text within images plays a crucial role in conveying information across various domains. Thus, extensive research in VQA has focused on text-rich images, such as documents~\cite{mathew2021docvqa, masry2022chartqa}, scene text~\cite{singh2019towards, mishra2019ocrvqa}, and digital interfaces~\cite{hsiao2022screenqa}, driving advances in Vision-Language Models (VLMs)~\cite{liu2023visual, wang2024qwen2, zhang2024llavar} designed to handle these diverse visual contexts. Recently, the field has progressed beyond basic text recognition, with new benchmarks~\cite{yue2024mmmupro, hao2025emma} emphasizing higher-order reasoning over textual content within images. Addressing these challenges necessitates tightly integrated cross-modal understanding, leveraging domain knowledge and multi-step reasoning that cannot be achieved by treating visual and linguistic elements in isolation.

However, low-resource languages including Korean lack benchmark suites even for basic text recognition, much less reasoning, impeding comprehensive evaluation and hindering model development across diverse domains (e.g., commerce, education) and image types (e.g., street signs, charts).
Although recent multilingual VQA benchmarks~\cite{tang2024mtvqa, sun2024parrot} have begun to address this disparity, they often struggle to provide sufficient coverage and depth for all languages. Existing Korean VQA datasets~\cite{ju2024varco, kim2025koffvqa} often rely on translated English questions and non-Korean images, or are limited in scale (e.g., fewer than 650 samples). 

To fill the underexplored evaluation gap for Korean text-rich VQA, we propose \textbf{KRETA}, a benchmark for \textbf{K}orean \textbf{R}eading and r\textbf{E}asoning in \textbf{T}ext-rich VQA \textbf{A}ttuned to diverse visual contexts. Specifically, Figure~\ref{fig:overview-of-kreta} (a) shows how KRETA is built upon a wide range of real-world Korean imagery, which we systematically categorized into 15 domains by referring to the Korean Standard Industrial Classification (KSIC)~\cite{ksic} and 26 image types widely used in prior works~\cite{yue2023mmmu, tang2024mtvqa}. Furthermore, we carefully design a dual-level reasoning framework inspired by the concepts of \textit{System 1} and \textit{System 2}~\cite{kahneman2011thinking}: \textit{System 1} assesses basic text recognition, while \textit{System 2} evaluates advanced capabilities such as domain-specific knowledge understanding, multi-step reasoning, and visual-based mathematical reasoning. KRETA comprises 2,577 samples, including 1,426 \textit{System 1} QA pairs and 1,151 \textit{System 2} QA pairs, and is, to the best of our knowledge, among the largest Korean text-rich VQA datasets currently available.

To ensure scalability and quality, we design a semi-automated VQA generation pipeline, as illustrated in Figure~\ref{fig:overview-of-kreta} (b). Unlike prior approaches~\cite{chen2024mj}, our method is specifically tailored for text-rich settings, centering on a refined, stepwise and multi-model decomposition that merges multiple VLM outputs to create high-quality structured captions for each image. This process is critical not only for capturing both textual and visual context, but also for minimizing hallucinations. Using these captions, we generate and evaluate QA candidates, synthesize hard negatives, and conduct final human refinement to ensure benchmark fidelity. We also release all prompts for question generation, as well as our seven evaluation metrics specifically designed for text-rich VQA, to support transparent adaptation and reproducibility.

Finally, our empirical analysis leveraging KRETA reveals that while VLMs demonstrate proficiency in basic Korean text recognition (System 1), a significant bottleneck remains for higher-order tasks requiring multi-step reasoning (System 2), particularly in open-source models. These models notably struggle with domain-specific knowledge and complex layouts, showing pronounced difficulty in areas like CSAT History and Marketing, as well as with image types such as banners and store signs. This underscores the need for targeted training on data encompassing Korean cultural and domain-specific knowledge, complex real-world layouts, and multi-step reasoning tasks. Our key contributions are threefold:


\begin{figure*}[!t]
  \centering 
  \includegraphics[width=0.93\textwidth]{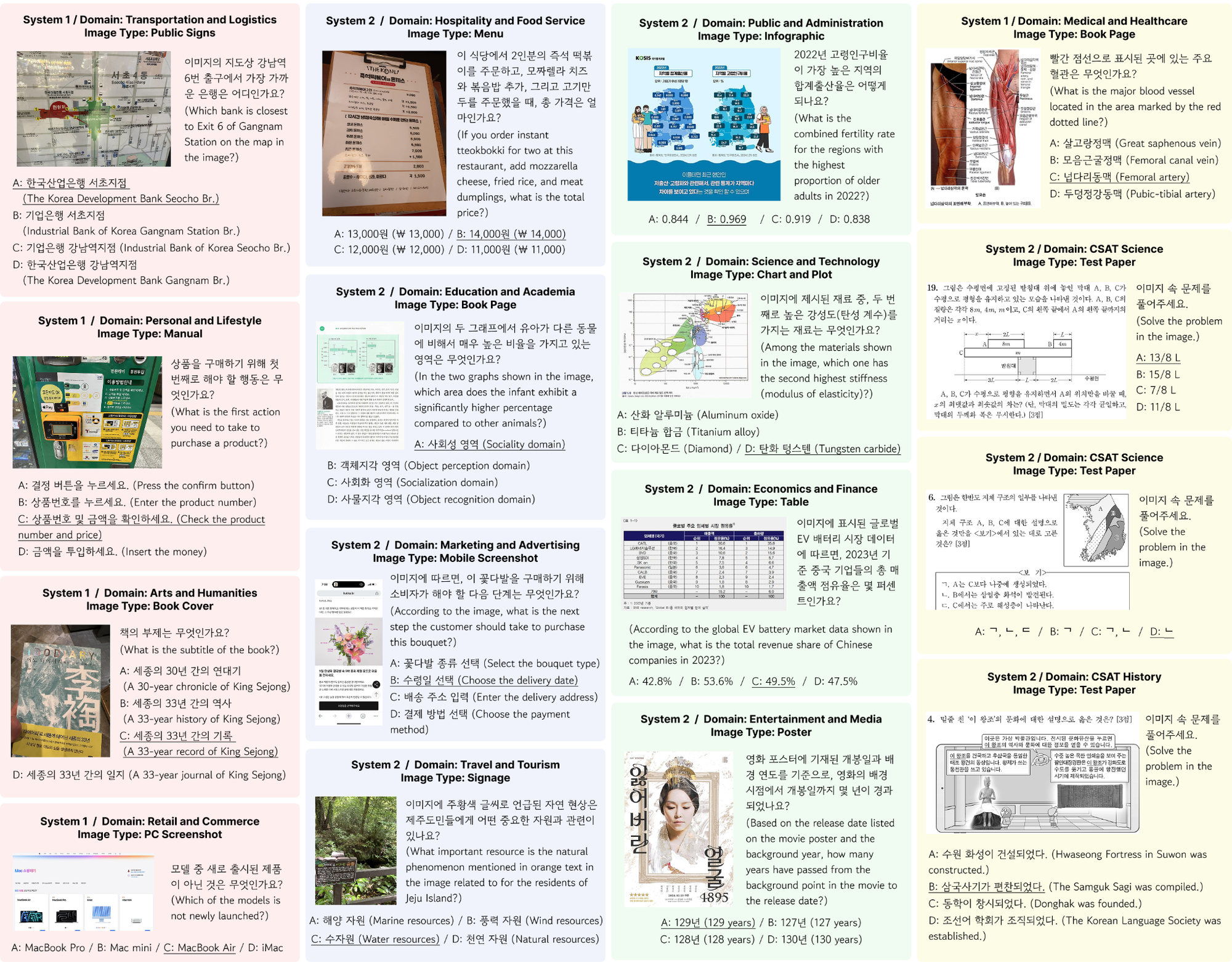}
  \caption{Examples from KRETA, showcasing diverse domains and image types categorized under System 1 and System 2. The model input consists of an image, a Korean question, and multiple-choice options.}
  \label{fig:example-of-kotextvqa}
\end{figure*}

\begin{enumerate}
\item \textbf{An in-depth and multi-faceted evaluation framework:} We adopt a dual-level reasoning framework, \textit{System 1} for basic understanding and \textit{System 2} for advanced reasoning, to provide an in-depth evaluation of VLM performance on text-rich images. Additionally, we adopt a multifaceted classification framework for images based on \textit{domain} and \textit{image type} to facilitate task-specific usage and evaluation in real-world industrial applications.

\item \textbf{A semi-automated VQA generation pipe-line:} We present a systematic and scalable pipeline optimized for text-rich VQA, featuring refined stepwise image decomposition and a seven-metric evaluation protocol to ensure data quality. To support adaptation to other low-resource languages, we release not only the dataset but also all prompts and code.

\item \textbf{A comprehensive text-rich VQA benchmark for Korean:} By integrating the above approaches, KRETA offers the first large-scale, high-quality benchmark to assess both basic and advanced reasoning of VLMs on real-world, text-rich Korean images spanning diverse domains and image types. 
\end{enumerate}

\section{Related Work}

\subsection{Vision-Language Models}

Recent advancements in VLMs~\cite{bai2025qwen2.5, abdin2024phi3, chen2024internvl2.5, wu2024deepseekvl2} have broadened their capabilities beyond traditional computer vision tasks, enabling contextual reasoning across visual domains and deeper language-vision integration. However, general-purpose VLMs often struggle with text-rich images, as they focus on holistic scene interpretation rather than precise text comprehension. To address this, text-centric VLMs such as LLaVAR~\cite{zhang2024llavar}, LLaVA-Read~\cite{zhang2024llavaread}, and TextSquare~\cite{tang2024textsquare} enhance reading abilities by refining text recognition and reasoning. While these models improve performance on text-rich tasks, they are still English-only, highlighting the need for multilingual VLMs.

\subsection{Text-Rich VQA Benchmarks}

General VQA benchmarks~\cite{lu2022learn,yue2023mmmu,yuan2023mmbench} evaluate broad reasoning skills. However, benchmarks dedicated to text-rich VQA remain scarce, especially outside English. Early work such as TextVQA~\cite{singh2019towards} and OCR-VQA~\cite{mishra2019ocrvqa} targets printed English text (e.g., billboards, book covers). 
Moving beyond English, MTVQA~\cite{tang2024mtvqa} provides multilingual annotations but is limited in scale, whereas MUST-VQA~\cite{vivoli2022mustvqa} expands data via automatic translation at the cost of language-specific nuance. xGQA~\cite{pfeiffer2022xgqa} also relies on machine translation with only a single difficulty tier, and SEA-VQA~\cite{urailertprasert2024seavqa} narrows its scope to Southeast-Asian heritage imagery.  
Meanwhile, KOFFVQA~\cite{kim2025koffvqa}, though rule-based, remains small and still merges reading with reasoning. Most text-oriented VQA benchmarks favor high-resource languages or depend on translated English datasets, with limited support for Korean~\cite{sun2024parrot,yue2024pangea}.

\section{KRETA Benchmark}

As shown in Table~\ref{tab:korean-vqa-datasets}, the KRETA benchmark is carefully designed to evaluate the ability of VLMs to understand and reason about Korean text appearing in images. Importantly, rather than relying on translated English resources, all images and QA pairs in KRETA were originally generated in Korean, ensuring natural language usage and cultural relevance. To the best of our knowledge, with 2,577 samples, it stands among the largest Korean text-rich VQA datasets to date. As illustrated in Figure~\ref{fig:example-of-kotextvqa}, KRETA features diverse visual contexts and requires advanced reasoning like domain-knowledge and multi-step cross-modal reasoning. The following subsections detail the dataset statistics and categorization, the data collection process, the semi-automated VQA generation pipeline, and the human annotation refinement process.

\subsection{Data Statistics and Categorization}
\label{ref:categorization}
Our benchmark consists of 2,577 samples, each annotated with corresponding QA pairs. Each image is categorized into one or both reasoning levels: \textit{System~1} (basic recognition and understanding) and \textit{System~2} (advanced reasoning). In total, the dataset includes 1,426 System~1 QA pairs and 1,151 System~2 QA pairs. Beyond the in-depth analysis provided by the reasoning-based categorization, we conduct a multi-faceted analysis of VLM performance by categorizing images along two additional dimensions: \emph{Domain} and \emph{Image Type}. The images cover 26 distinct types across 15 domains.

\begin{table}
  \centering
  \resizebox{0.48\textwidth}{!}{%
  \begin{tabular}{l|c|r|c|c|c}
    \toprule
    \textbf{Benchmark} & \makecell{\textbf{Image} \\ \textbf{Source}} & \textbf{Samples} & \makecell{\textbf{Text-Centric} \\ \textbf{Reasoning}} & \textbf{Forms} & \makecell{\textbf{Image} \\ \textbf{Type}} \\
    \midrule
    K-MMB~\cite{ju2024varco} & En & 4,329 & - & MC & General \\
    K-SEED~\cite{ju2024varco} & En & 2,971 & - & MC & General \\
    K-MMSTAR~\cite{ju2024varco} & En & 1,500 & - & MC & General \\
    K-LLaVA-W~\cite{ju2024varco} & En & 60 & - & Open & General \\
    \midrule
    K-Viscuit~\cite{baek2024evaluating} & Ko & 657 & - & MC & General \\
    K-DTCBench~\cite{ju2024varco} & Ko & 240 & \checkmark & MC & Document \\
    MTVQA-ko~\cite{tang2024mtvqa} & Ko & 558 & - & Short & Multi-text \\
    KOFFVQA~\cite{kim2025koffvqa} & Ko & 275 & \checkmark & Open & Multi-text \\
    \midrule\midrule
    \rowcolor{cyan!10}
    KRETA (Ours) & Ko & 2,577 & \checkmark & MC & Multi-text \\
    \bottomrule
  \end{tabular}
  }
  \caption{\label{tab:korean-vqa-datasets} Comparison of Korean VQA Benchmarks. The Image Source column indicates native Korean images (Ko) or English-translated ones (En). Text-Centric Reasoning indicates whether the benchmark focuses on reasoning over text in images. Forms lists the answer type open-ended (Open), short-answer (Short), or multiple-choice (MC). Image Type categorizes images as General (non text-centric), Document (structured layouts), or Multi-text (diverse text-rich contexts).}
\end{table}

\paragraph{System~1 vs. System~2}\label{ref:system1/2}
To assess challenges in visual text understanding and provide a comprehensive evaluation, we adopt a two-tiered cognitive framework~\cite{kahneman2011thinking, yu2024distilling} that distinguishes basic recognition (System 1, fast thinking) from advanced reasoning (System 2, slow thinking). 
System~1 relies on intuitive and automatic recognition, requiring direct text extraction and straightforward interpretation. In contrast, System~2 demands advanced reasoning, such as contextual understanding, multi-step decision-making, numerical reasoning, and integration of external knowledge when necessary. 

\paragraph{Domain}\label{ref:domain}
To ensure that our domain classification aligns with real-world industrial applications, we refer to the Korean Standard Industrial Classification (KSIC)~\cite{ksic} framework. We adapt this framework to suit our image data analysis, following a structured approach similar to MMMU~\cite{yue2023mmmu}. We define 13 primary domains: Public \& Administration (Gov.), Economics \& Finance (Econ.), Marketing \& Advertising (Mktg.), Retail \& Commerce (Comm.), Education \& Academia (Edu.), Medical \& Healthcare (Med.), Science \& Technology (Tech.), Arts \& Humanities (Arts.), Transportation \& Logistics (Transp.), Travel \& Tourism (Tour.), Hospitality \& Food Service (FnB.), Entertainment \& Media (Ent.), and Personal \& Lifestyle (Life.).

In addition, we incorporate CSAT (College Scholastic Ability Test) Science (Sci.) and History (Hist.) as separate domains. Unlike other domains generated by our semi-automated pipeline, CSAT questions were directly adapted from official examination materials. For each item, we crop the image region containing the question stem and its associated visual context, then supply the multiple-choice options to the model as text.

\paragraph{Image Type}  
Images are categorized based on their inherent visual structures and the way they convey information. To systematically analyze VLM performance across different visual formats, we classify all images into 26 distinct types, ranging from highly structured (e.g., tables, receipts) to visually complex (e.g., posters, PC screenshots). Detailed definitions for all 26 image types are provided in Appendix~\ref{app:image-types}.

\subsection{Data Collection}

For this study, we compile images from copyright-free online repositories and our own field photography. To ensure balanced coverage of real-world scenarios, we identify domain imbalances and add samples in underrepresented categories. We source data from government publications, posters, free image databases, administrative documents, statistical reports from public agencies, and publicly available Korean mock exams including the CSAT.

\begin{table}
  \centering
  \resizebox{0.48\textwidth}{!}{%
  \begin{tabular}{l|c|c|c>{\columncolor{cyan!10}}c}
    \toprule
    \textbf{Model} & \textbf{Size} & \makecell{\textbf{Overall} \\ (2,577)} & \makecell{\textbf{System 1} \\ (1,426)} & \makecell{\textbf{System 2} \\ (1,151)} \\
    \midrule
    \multicolumn{5}{l}{\textbf{\textit{Closed}}} \\ 
    \midrule
    GPT-4o~\cite{gpt4o} & - & 84.6 & 95.9 & \textbf{70.5} \\
    GPT-4o-mini~\cite{gpt4o} & - & 73.3 & 88.7 & 54.1 \\
    Gemini-2.0-flash~\cite{gemini2} & - & \textbf{85.4} & \textbf{98.0} & 69.8 \\
    Claude-3.5-Sonnet~\cite{anthropic2024claude} & - & 80.5 & 93.4 & 64.5 \\
    \midrule
    \multicolumn{5}{l}{\textbf{\textit{Open-source}}} \\ 
    \midrule
    LLaVA-OneVision~\cite{li2024llavaov} & 0.5B & 42.3 & 49.6 & 33.3 \\
    Deepseek-VL2-tiny~\cite{wu2024deepseekvl2} & 1B & 48.8 & 60.8 & 34.0 \\
    Deepseek-VL2-small~\cite{wu2024deepseekvl2} & 2.8B & 53.3 & 67.3 & 36.1 \\
    Qwen2.5-VL~\cite{wang2024qwen2} & 3B & \textbf{71.8} & \textbf{94.2} & 43.9 \\
    Ovis1.6-Llama3.2~\cite{lu2024ovis} & 3B & 52.2 & 62.8 & 39.1 \\
    InternVL2.5~\cite{chen2024internvl2.5} & 4B & 70.7 & 90.7 & \textbf{45.9} \\
    Phi-3.5-Vision~\cite{abdin2024phi3} & 4.2B & 42.6 & 52.2 & 30.8 \\
    \midrule
    LLaVA-OneVision~\cite{li2024llavaov} & 7B & 54.0 & 65.1 & 40.1 \\
    Qwen2.5-VL~\cite{wang2024qwen2} & 7B & 68.5 & \textbf{94.5} & 36.1 \\
    InternVL2.5~\cite{chen2024internvl2.5} & 8B & 70.8 & 89.8 & 47.3 \\
    MiniCPM-V-2.6~\cite{yao2024minicpm} & 8B & 41.0 & 50.4 & 29.4 \\
    MiniCPM-o-2.6~\cite{yao2024minicpm} & 8B & 64.3 & 84.1 & 39.9 \\
    Ovis1.6-Gemma2~\cite{lu2024ovis} & 9B & 58.4 & 68.9 & 45.4 \\
    VARCO-VISION~\cite{ju2024varco} & 14B & \textbf{72.3} & 90.9 & \textbf{49.3} \\
    \bottomrule
  \end{tabular}
  }
  \caption{\label{tab:system-results}
Evaluation results of closed and open-source VLMs on the KRETA, highlighting performance under the System~1 and System~2 framework. As marked in color, models struggle with System~2 reasoning tasks.
}
\label{tab:system-comparison}
\end{table}

\begin{table*}
\centering
\resizebox{\textwidth}{!}{%
\begin{tabular}{l|c|c|ccccccccccccc|cc}
\toprule
\textbf{Model}
& \textbf{Size}
& \makecell{\textbf{Overall}\\(2,577)}
& \makecell{\textbf{Gov.}\\(245)}
& \makecell{\textbf{Econ.}\\(104)}
& \makecell{\textbf{Mktg.}\\(145)}
& \makecell{\textbf{Comm.}\\(154)}
& \makecell{\textbf{Edu.}\\(215)}
& \makecell{\textbf{Med.}\\(90)}
& \makecell{\textbf{Tech.}\\(92)}
& \makecell{\textbf{Arts.}\\(83)}
& \makecell{\textbf{Transp.}\\(167)}
& \makecell{\textbf{Tour.}\\(108)}
& \makecell{\textbf{FnB.}\\(264)}
& \makecell{\textbf{Ent.}\\(168)}
& \makecell{\textbf{Life.}\\(204)}
& \makecell{\textbf{\textit{Sci.}}\\(478)}
& \makecell{\textbf{\textit{Hist.}}\\(60)} \\
\midrule
\multicolumn{18}{l}{\textbf{\textit{Closed}}} \\ 
\midrule
GPT-4o~\cite{gpt4o} & -
& 84.6  & 93.5  & 92.3  & 97.2  & 90.3  & \textbf{96.7}  & 91.1  & \textbf{96.7}  & \textbf{100.0} & 84.4  & 93.5  & \textbf{93.6}  & \textbf{97.0}  & 95.1  & \textbf{44.1}  & \textbf{93.3} \\ 
GPT-4o-mini~\cite{gpt4o} & -
& 73.3  & 82.4  & 82.7  & 85.5  & 84.4  & 87.4  & 83.3  & 80.4  & 89.2  
& 80.2  & 84.3  & 81.4  & 86.3  & 87.3  & 30.3  & 45.0 \\
Gemini-2.0-flash~\cite{gemini2} & -
& \textbf{85.4}  & \textbf{95.1}  & \textbf{95.2}  & \textbf{99.3}  & \textbf{96.1}  & \textbf{96.7} & \textbf{92.2}  & 93.5  & 98.8  
& \textbf{90.4}  & \textbf{98.1}  & 93.2  & 95.2  & \textbf{96.6}  & \textbf{44.1}  & 78.3 \\
Claude-3.5-Sonnet~\cite{anthropic2024claude} & -
& 80.5  & 93.5  & 91.3  & 92.4  & 87.0  & 93.0  & 91.1  & 87.0  & 91.6  
& 84.4  & 94.4  & 89.8  & 92.3  & 92.2  & 37.4  & 70.0 \\
\midrule
\multicolumn{18}{l}{\textbf{\textit{Open-source}}} \\ 
\midrule
LLaVA-OneVision~\cite{li2024llavaov} & 0.5B
& 42.3  & 51.8  & 48.1  & 47.6  & 44.8  & 39.5  & 50.0  & 44.6  & 40.9  
& 49.7  & 51.9  & 41.7  & 44.6  & 46.1  & 28.0  & 31.7 \\
Deepseek-VL2-tiny~\cite{wu2024deepseekvl2} & 1B
& 48.8  & 57.1  & 55.8  & 63.4  & 58.4  & 51.2  & 57.8  & 57.6  & 45.8  
& 54.5  & 58.3  & 43.9  & 47.0  & 54.4  & 30.5  & 31.7 \\
Deepseek-VL2-small~\cite{wu2024deepseekvl2} & 2.8B
& 53.3  & 61.6  & 63.5  & 66.9  & 63.0  & 57.2  & 64.4  & 68.5  & 50.6  
& 59.9  & 63.0  & 48.9  & 56.0  & 57.4  & 30.8  & 36.7 \\
Qwen2.5-VL~\cite{wang2024qwen2} & 3B
& \textbf{71.8}  & 81.6  & \textbf{76.9}  & 85.5  & 77.9  & \textbf{87.4}  & \textbf{80.0}  & \textbf{79.3}  & \textbf{85.5} & \textbf{75.4}  & \textbf{84.3} & \textbf{76.9}  & \textbf{87.5}  & 83.3  & \textbf{33.9}  & 36.7 \\
Ovis1.6-Llama3.2~\cite{lu2024ovis} & 3B
& 52.2  & 64.5  & 69.2  & 60.7  & 57.1  & 55.8  & 54.4  & 62.0  & 51.8  
& 60.5  & 61.1  & 56.8  & 52.4  & 49.5  & 30.5  & 31.7 \\
InternVL2.5~\cite{chen2024internvl2.5} & 4B
& 70.7  & \textbf{82.0}  & \textbf{76.9}  & \textbf{87.6}  & \textbf{83.1}  & 83.7  & 78.9  & \textbf{79.3}  & 79.5 & 75.4  & 77.8  & 69.3  & 81.0  & \textbf{86.3}  & \textbf{33.9} & \textbf{46.7} \\
Phi-3.5-Vision~\cite{abdin2024phi3} & 4.2B
& 42.6  & 53.5  & 55.8  & 40.0  & 49.4  & 43.3  & 40.0  & 53.3  & 50.6  
& 44.3  & 46.3  & 42.8  & 43.5  & 44.6  & 27.6  & 36.7 \\ 
\midrule
LLaVA-OneVision~\cite{li2024llavaov} & 7B
& 54.0  & 64.1  & 63.5  & 63.4  & 63.6  & 58.6  & 55.6  & 64.1  & 45.8  
& 68.3  & 65.7  & 55.3  & 55.4  & 55.9  & 30.8  & 33.3 \\
Qwen2.5-VL~\cite{wang2024qwen2} & 7B
& 68.5  & 80.0  & 77.9  & \textbf{85.5}  & 81.2  & \textbf{87.4}  & 76.7  & 75.0  & \textbf{89.2}  
& 77.8  & 82.4  & \textbf{77.7}  & \textbf{86.3}  & \textbf{85.8}  & 15.1  & 36.7 \\
InternVL2.5~\cite{chen2024internvl2.5} & 8B
& 70.8  & \textbf{81.6}  & 76.9  & \textbf{85.5}  & 81.8  & 83.7  & 81.1  & 77.2  & 78.3  
& 76.0  & \textbf{83.3}  & 74.2  & 78.6  & \textbf{85.8}  & \textbf{34.1}  & \textbf{38.3} \\
MiniCPM-V-2.6~\cite{yao2024minicpm} & 8B
& 41.0  & 50.2  & 54.8  & 50.3  & 53.2  & 44.7  & 41.1  & 52.2  & 33.7  
& 43.7  & 48.1  & 43.6  & 45.8  & 46.1  & 18.2  & 25.0 \\
MiniCPM-o-2.6~\cite{yao2024minicpm} & 8B
& 64.3  & 75.9  & 83.7  & 79.3  & 75.9  & 76.7  & 65.6  & 75.0  & 73.5  
& 69.5  & 79.6  & 67.8  & 77.4  & 74.0  & 25.5  & 25.0 \\
Ovis1.6-Gemma2~\cite{lu2024ovis} & 9B
& 58.4  & 64.1  & 69.2  & 71.0  & 72.7  & 60.9  & 71.1  & 67.4  & 53.0  
& 68.9  & 75.9  & 65.2  & 58.9  & 63.2  & 30.5  & 28.3 \\
VARCO-VISION~\cite{ju2024varco} & 14B
& \textbf{72.3}  & \textbf{81.6}  & \textbf{87.5}  & 83.4  & \textbf{83.1}  & 84.2  & \textbf{86.7}  & \textbf{84.8}  & 79.5  
& \textbf{82.6}  & \textbf{83.3}  & 76.1  & 81.5  & 85.3  & 33.7  & 31.7 \\
\bottomrule
\end{tabular}
}
\caption{\label{tab:domain-results}
Evaluation results for closed and open-source VLMs on KRETA across 15 domains. 
}
\label{tab:domain-abbrev}
\end{table*}

\subsection{Semi-Automated VQA Generation Pipeline}

\paragraph{Step 1: Stepwise Image Decomposition}
In this step, we refine the dataset by filtering out low-quality images. Images with a shortest side of 384 pixels or less are discarded to ensure text readability. To further ensure meaningful textual content, we use \href{https://github.com/PaddlePaddle/PaddleOCR}{PaddleOCR}\footnote{\url{https://github.com/PaddlePaddle/PaddleOCR}} to exclude images with fewer than 10 or more than 1,000 Korean characters. 

Following the filtering process, multiple VLMs independently extract both textual and non-textual elements from each image. By default, we employ two foundation models, GPT-4o-mini~\cite{gpt4o} and Gemini-2.0-flash~\cite{gemini2}, and merge their outputs to maximize extraction thoroughness while minimizing hallucinations. The structured decomposition process first analyzes non-textual visual attributes such as the overall scene, document layout, key objects, and background details. It then examines the structural and semantic relationships between text and visual components before finally extracting and structuring all textual content. This approach preserves contextual links between visual and textual elements, yielding higher-quality outputs than direct OCR alone.

\paragraph{Step 2: QA Candidates Generation}  

Using the structured captions from Step 1, this step simultaneously generates question-answer candidates via LLMs. QA generation follows the System 1 and System 2 framework, with prompts specifically designed to assess different levels of visual text understanding and reasoning. For System 1, we use GPT-4o-mini~\cite{gpt4o} and Gemini-2.0-flash~\cite{gemini2} to generate two candidates each. For System 2, we employ o1-mini~\cite{o1-mini} and Gemini-2.0-flash~\cite{gemini2} to leverage their strong reasoning performance. The pipeline offers flexible control over the choice of model and the number of QA candidates generated.  

Independently, the classification step assigns each image to its appropriate domain and image type as defined in Section~\ref{ref:categorization}, based on the structured captions from Step 1.

\paragraph{Step 3: QA Evaluation and Voting} 

In this step, multiple VLMs (by default GPT-4o-mini~\cite{gpt4o} and Gemini-2.0-flash~\cite{gemini2}) evaluate the generated QA candidates to determine the highest-quality question-answer pair for each image. Drawing inspiration from prior LLM evaluation research~\cite{zheng2023judging, fu2024qgeval}, the process employs a set of predefined criteria to systematically assess candidate quality.

For System 1 candidates, we use five metrics (Text Utilization, Clarity, Correctness, Naturalness, and Alignment) to ensure textual content accuracy and coherence. For System 2 candidates, two additional metrics (Complexity and Coherence) account for multi-step reasoning and logical inference. Each VLM assigns a score from 0 to 5 for each metric, and we use the aggregated scores to rank the candidates. A voting mechanism then selects the highest-ranked QA pair across all VLMs.

\paragraph{Step 4: Hard Negatives Generation}  
After selecting the final QA pair, an LLM generates three hard negative options that resemble the correct answer while remaining distinct in meaning. These options follow the correct answer's structure and context, making the multiple-choice format more challenging. 

\paragraph{Human Annotation Refinement}  

The final QA pairs undergo a thorough human review based on the same evaluation criteria as Step 3. We adjust or remove questions that can be answered solely from text without image context (Text Utilization); verify that each QA pair aligns with the image's original intent (Alignment); confirm that System 2 questions require at least one inferential step to avoid overly simple QA (Complexity); and review language, grammar, and factual content (Naturalness, Correctness, and Clarity).

\section{Empirical Analysis}
We leverage VLMEvalKit~\cite{duan2024vlmevalkit}, an open-source evaluation toolkit designed to facilitate the assessment of VLMs, including both proprietary APIs and open-source models. We adopt the multiple-choice system prompt from MMMU-Pro~\cite{yue2024mmmupro}. The prompt instructs the model as follows: \textit{Please select the correct answer from the options above. The last line of your response should follow the format: `Answer: LETTER' (without quotes), where LETTER corresponds to one of the provided options.}

\subsection{Performance across System 1 vs. System 2}
Table~\ref{tab:system-comparison} presents the performance breakdown between System 1 and System 2. Across both open-source and closed models, System 1 accuracy is significantly higher, indicating that most models handle text recognition and simple contextual understanding well. Notably, Gemini-2.0-flash~\cite{gemini2} achieves 98.0\% on System 1, reflecting near-perfect perception.

However, System 2 results reveal substantial performance drops, particularly in open-source models. Qwen2.5-VL-7B~\cite{wang2024qwen2} falls from 94.5\% in System 1 to 36.1\% in System 2, and Deepseek-VL2-small~\cite{wu2024deepseekvl2} drops from 67.3\% to 36.1\%. GPT-4o~\cite{gpt4o} retains a relatively stronger System 2 performance at 70.5\%, yet this value remains suboptimal. Both open-source and closed models struggle in System 2, as effective reasoning requires sequential integration of multiple visual and textual cues, a capability that is still underdeveloped. These challenges are compounded by the low-resource nature of Korean pretraining, and by gaps in domain-specific and cultural knowledge, since models have limited exposure to Korean-contextualized data during training.

\subsection{Performance across Domain}

Table~\ref{tab:domain-abbrev} compares closed and open-source model performance across 15 domains. Among closed models, Gemini-2.0-flash~\cite{gemini2} achieves the highest overall score (85.4\%), followed by GPT-4o (84.6\%). Notably, GPT-4o excels in the CSAT History domain with 93.3\%, suggesting strong historical and cultural reasoning. Gemini-2.0-flash's consistently high performance across domains further reflects its robust text recognition and contextual comprehension on real-world images.

Open-source models exhibit a broad range of performance on KRETA, with overall scores varying from 42.3\% (LLaVA-OneVision~\cite{li2024llavaov}, 0.5B) to 72.3\% (VARCO-VISION~\cite{ju2024varco}, 14B). The strongest performers are Qwen2.5-VL~\cite{wang2024qwen2}, InternVL2.5~\cite{chen2024internvl2.5}, and VARCO-VISION, each scoring in the low 70\% range. Notably, Qwen2.5-VL (7B) maintains high accuracy in practical domains such as Marketing (85.5\%) yet plunges to 15.1\% in CSAT Science. Applying Chain-of-Thought prompting (Section~\ref{sec:CoT}) substantially boosts Qwen2.5-VL's System 2 performance overall, underscoring its original deficiency in multi-step reasoning and external knowledge integration. Taken together, Table~\ref{tab:domain-abbrev} reveals significant variability among open-source models in both overall and domain-specific metrics, highlighting the need to carefully consider model size, architecture, and domain alignment when selecting a model for a given application.

\begin{figure}
  \centering
  \includegraphics[width=0.48\textwidth]{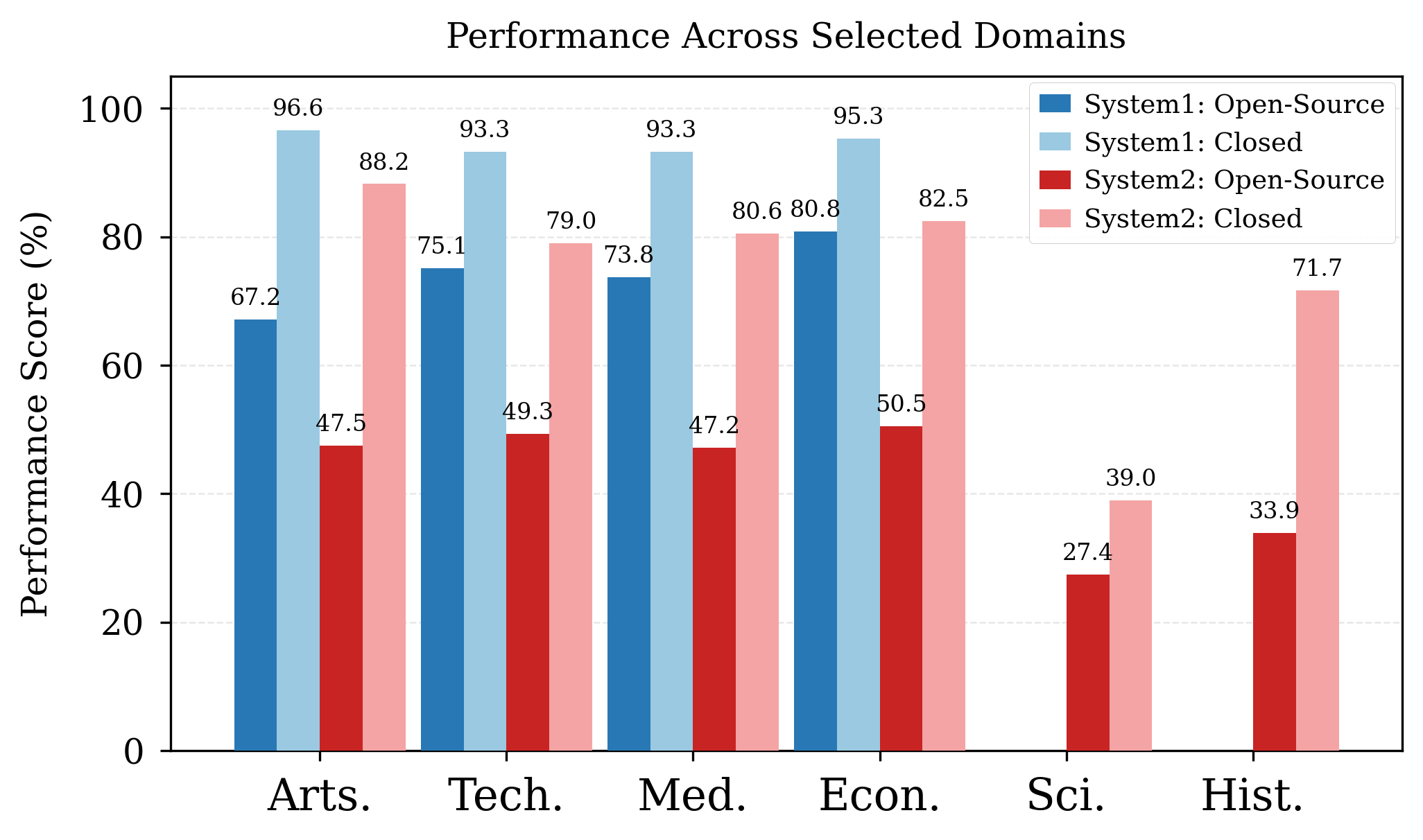}
  \vspace{-0.5cm}
  \caption{Comparison of open-source and closed models across different domains on KRETA. Bars show the average scores of closed and open-source models separately for System~1 and System~2 in each domain.
}
  \label{fig:open-closed-comparison}
\end{figure}

Figure~\ref{fig:open-closed-comparison} illustrates the System~1 and System~2 performance gap between closed and open-source models across different domains on KRETA. The disparity is particularly pronounced in System~2 tasks, where closed models outperform open-source counterparts by up to 40.7 percentage points in \textit{Arts \& Humanities}, reflecting stronger cultural understanding. Meanwhile, the \textit{Science \& Technology} domain shows a relatively smaller System~2 gap of 29.7 percentage points, suggesting more consistent handling of technical content. In the CSAT domains, gaps of 11.6 points in Science and 37.8 points in History further underscore the role of background knowledge. These findings suggest that open-source models need targeted domain-specific training, particularly in culturally and historically rich areas, to close the reasoning gap with closed models.

\begin{table}
  \centering
  \resizebox{0.48\textwidth}{!}{%
  \begin{tabular}{l|cc|cc|cc|cc}
    \toprule
    \textbf{Image Type} & \multicolumn{2}{c|}{\textbf{Closed}} & \multicolumn{2}{c|}{\textbf{Open}} & \multicolumn{2}{c|}{\textbf{Sys1 - Sys2}} & \multicolumn{2}{c}{\textbf{Closed - Open}} \\
    \cmidrule(r){2-3} \cmidrule(r){4-5} \cmidrule(r){6-7} \cmidrule(r){8-9}
     & \textbf{Sys1} & \textbf{Sys2} & \textbf{Sys1} & \textbf{Sys2} & \textbf{Closed} & \textbf{Open} & \textbf{Sys1} & \textbf{Sys2} \\
    \midrule
    \multicolumn{4}{l}{\textbf{\textit{Document}}} \\ 
    \midrule
    Chart and Plot   & 94.9 & 86.7 & 79.3 & 48.2 & 8.2  & 31.1 & 15.6 & 38.5 \\
    Table            & 91.0 & 75.0 & 70.9 & 42.3 & \textbf{16.0} & 28.6 & 20.1 & 32.7 \\
    Infographic      & 95.4 & 81.3 & \textbf{80.0} & 44.1 & 14.1 & \textbf{35.9} & 15.4 & 37.2 \\
    Slides           & 96.4 & \textbf{95.0} & 73.0 & \textbf{61.3} & 1.4  & 11.7 & 23.4 & 33.7 \\
    Book Cover       & 95.4 & 91.0 & 69.0 & 52.0 & 4.4  & 17.0 & \textbf{26.4} & \textbf{39.0} \\
    Product Detail   & 94.3 & 87.5 & 78.6 & 51.5 & 6.8  & 27.1 & 15.7 & 36.0 \\
    Poster           & 94.6 & 87.3 & 73.8 & 54.0 & 7.3  & 19.8 & 20.8 & 33.3 \\
    Mobile Screen    & \textbf{97.2} & 90.7 & 76.9 & 54.9 & 6.5  & 22.0 & 20.3 & 35.8 \\
    PC Screen        & 94.8 & 83.6 & 74.8 & 50.1 & 11.2 & 24.7 & 20.0 & 33.5 \\
    \midrule
    \multicolumn{9}{l}{\textbf{\textit{Scene Text}}} \\ 
    \midrule
    Street Signs     & 87.0 & \textbf{93.1} & 75.9 & \textbf{59.3} & -6.1 & 16.6 & 11.1 & 33.8 \\
    Public Signs     & 88.6 & 69.4 & 71.2 & 42.0 & 19.2 & 29.2 & 17.4 & 27.4 \\
    Store Sign       & 91.4 & 85.3 & 70.6 & 42.0 & 6.1  & 28.6 & 20.8 & 43.3 \\
    Banner           & 94.6 & 91.1 & 78.2 & 46.2 & 3.5  & \textbf{32.0} & 16.4 & \textbf{44.9} \\
    Signage          & \textbf{94.7} & 85.9 & \textbf{78.5} & 54.3 & 8.8  & 24.2 & 16.2 & 31.6 \\
    Menu             & 91.9 & 79.9 & 69.5 & 40.3 & 12.0 & 29.2 & \textbf{22.4} & 39.6 \\
    Manual           & 91.2 & 71.1 & 73.2 & 42.1 & \textbf{20.1} & 31.1 & 18.0 & 29.0 \\
    \bottomrule
  \end{tabular}%
  }

  \caption{\label{tab:image-type-comparison}
    Performance comparison across image types for closed and open-source models, showing differences across System~1, System~2, and model categories. Only image types with at least 50 VQA pairs are presented.
  }
\label{tab:type-comparison}
\end{table}

\subsection{Performance across Image Type}  

Table~\ref{tab:image-type-comparison} presents the performance of closed and open-source models across different image types, highlighting key trends in System~1 and System~2 tasks. Performance varies significantly by image type, reflecting distinct model capabilities. Document-based images such as tables and infographics achieve high accuracy for closed-source models in System~1 (91.0\%, 95.4\%) and retain relatively strong performance in System~2 (75.0\%, 81.3\%). In contrast, open-source models fall to 42.3\% on Tables and 44.1\% on Infographics in System~2. Notably, Book Covers exhibit the largest closed-open gap: 26.4 points in System~1 and 39.0 points in System~2, likely due to their complex typography and mixed visual elements.

Scene-text images present different challenges. Street Signs show a rare pattern for closed models, with System~2 accuracy (93.1\%) exceeding System~1 (87.0\%), possibly because motion blur or low resolution impairs simple text extraction while System~2 can leverage broader context. In contrast, open-source models perform particularly poorly on banners and store signs, where the System~2 gap reaches 44.9 points and 43.3 points, respectively, indicating difficulties with diverse fonts, occlusions, and unconventional layouts common in real-world signage. These findings highlight the varying complexity of image types and underscore the need for targeted improvements in both structured-text processing and robust scene-text understanding.

\subsection{Performance across Closed vs. Open-source}

Overall, closed-source models outperform open-source counterparts by an average of 24.2 percentage points in overall score, with the System 2 reasoning gap reaching as high as 44.4 percentage points, revealing a pronounced “reasoning bottleneck” in open models (Table~\ref{tab:system-comparison}). 
Domain analysis shows a relatively modest closed-open gap of 23.7 percentage points in the \textit{Science \& Technology} domain, but this difference widens to 40.7 percentage points in CSAT History, highlighting the closed models' superior ability to integrate background knowledge and cultural context (Table~\ref{tab:domain-abbrev}). 
Similarly, across image types, from structured documents such as tables and infographics to cluttered, unstructured layouts such as banners and signage, the transition from closed to open models yields comparable performance declines, underscoring open models' limited versatility in handling diverse visual-textual presentations (Table~\ref{tab:type-comparison}).

\begin{figure}
  \centering
  \includegraphics[width=0.48\textwidth]{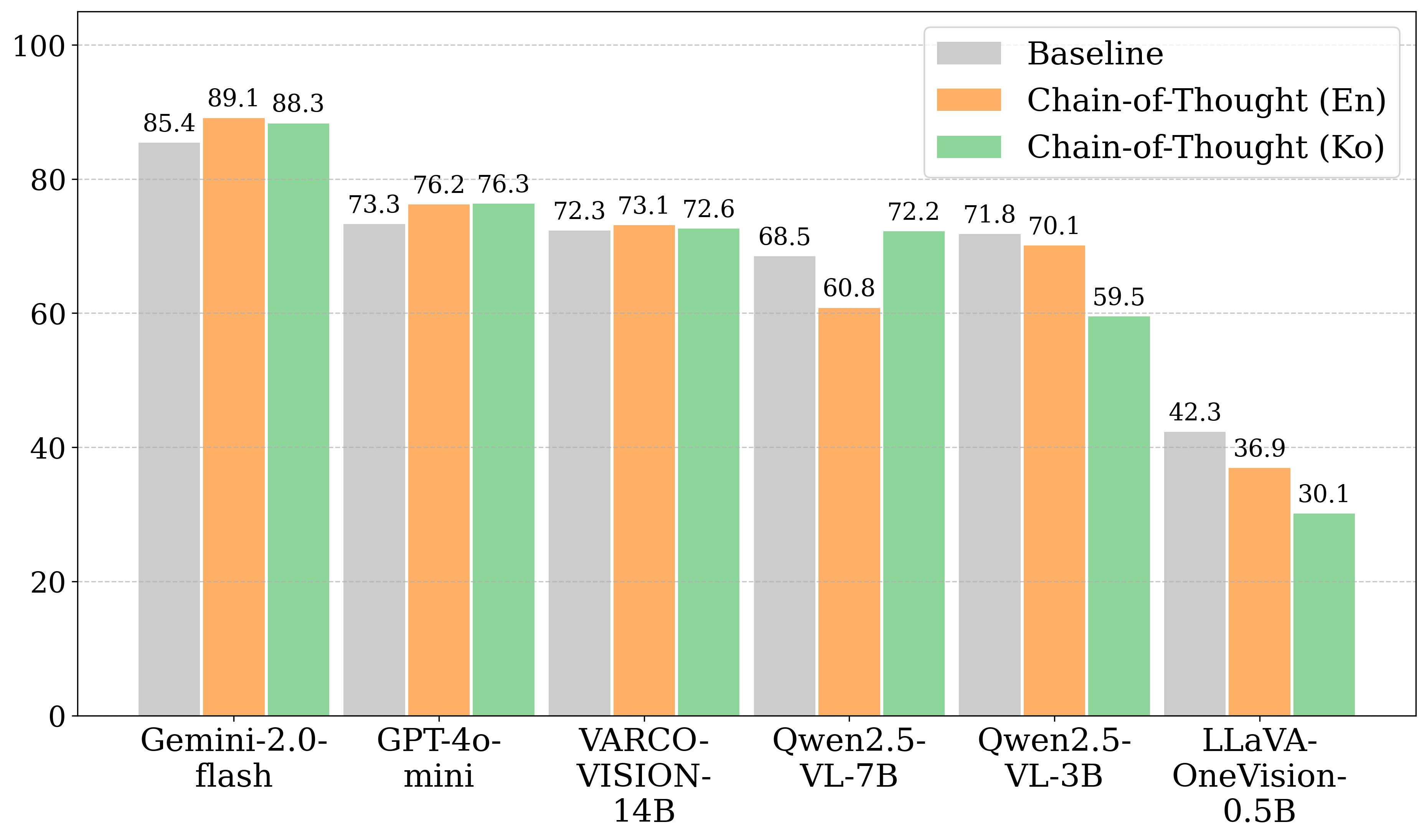}
  \caption{Comparison of two closed and four open-source models of varying sizes on KRETA. The figure shows performance differences across three prompts: Baseline, Chain-of-Thought in English and Korean.}
  \label{fig:cot}
\end{figure}

\subsection{Performance across Model Size}

Table~\ref{tab:domain-abbrev} demonstrates a clear positive correlation between model capacity and overall performance: Deepseek-VL2 improves from 48.8 at 1 B parameters (tiny) to 53.3 at 2.8 B (small), and LLaVA-OneVision rises from 42.3 at 0.5 B to 54.0 at 7 B. These results confirm that, for a given architecture, increasing model size generally yields gains in both aggregate accuracy and domain-specific metrics. An exception to this trend is observed with Qwen2.5-VL, where the 3 B variant (71.8) outperforms the 7 B variant (68.5). This anomaly suggests that the addition of further multilingual data during scaling may have diluted the model's Korean-centric knowledge and reasoning abilities. Consequently, when enlarging multilingual VLMs, it is essential to preserve the proportion of low-resource language data and to apply domain-adaptive fine-tuning to sustain performance on language-specific and culturally nuanced tasks.

\section{Discussion}

\paragraph{Chain-of-Thought (CoT)}\label{sec:CoT}

We evaluate the impact of Chain of Thought (CoT) prompting on model performance, following the approach demonstrated in MMMU-Pro~\cite{yue2024mmmupro}. Figure~\ref{fig:cot} reveals a pronounced gap between closed and open-source models in both baseline scores and CoT improvements.

Closed models benefit consistently. For instance, Gemini 2.0-flash improves by 3.7 points with English CoT and by 2.9 points with Korean CoT, indicating robust instruction following and structured reasoning. Mid-size open-source models exhibit language-dependent effects. Qwen2.5-VL-7B declines by 7.7 points with English CoT but improves by 3.7 points with Korean CoT, suggesting sensitivity to prompt language and potential for language-specific optimization. Lightweight open-source models suffer performance degradation under CoT prompting. LLaVA-OneVision-0.5B drops by 5.4 points under English CoT and by 12.2 points under Korean CoT. These declines suggest that excessive reasoning instructions overwhelm models with limited capacity.

Overall, these results demonstrate that CoT prompting enhances performance only when a model possesses adequate reasoning capacity and instruction-following ability, and may become detrimental otherwise.

\section{Conclusion}

In this paper, we present KRETA, a comprehensive benchmark for evaluating VLMs on Korean text-rich images. KRETA adopts a dual-level reasoning framework for both basic recognition and advanced inference, and employs an industry-aligned domain and image-type taxonomy spanning 15 domains and 26 image formats. By relying exclusively on native Korean imagery and questions, our benchmark extends beyond previous Korean or multilingual VQA sets that have been confined to document-only tasks or machine-translated content. Our training-free, semi-automated pipeline combines structured image decomposition with cross-validation by two foundation models and a novel multi-metric evaluation protocol. Our experimental results underscore the need for domain-adaptive fine-tuning, the careful preservation of low-resource language data balance during scaling, and the integration of stronger reasoning mechanisms. We hope that our VQA generation pipeline will be readily transferable to other low-resource languages, laying the groundwork for culturally and linguistically tailored VLMs.

\newpage 

\section*{Limitations}
\label{sec:limitation}

While we provide a comprehensive evaluation of Korean text-rich VQA, several limitations suggest directions for future work. First, KRETA is confined to single-image, multiple-choice question answering. Extending the benchmark to include multi-image or video-based scenarios, and to incorporate high-level comprehension tasks (e.g. section-to-section verification, information synthesis, document summarization, open-ended generation), would yield a more complete assessment of vision-language capabilities.

Second, the System 2 category conflates sequential deduction, integration of external information and cross-referential contextual analysis into a single classification. Developing a more fine-grained taxonomy to distinguish these reasoning functions would expose specific model weaknesses and support targeted improvements. In this work, we prioritized the creation of a scalable, high-quality unified benchmark spanning 15 domains and 26 image formats, establishing a solid foundation for Korean text-rich VQA while leaving finer reasoning taxonomies and additional task formats to future extensions.

Lastly, Chain-of-Thought (CoT) prompting has been shown to improve performance on the System 2 benchmark, particularly for closed-source models that consistently gain from both English and Korean CoT variants, but additional strategies and prompt formulations remain unexplored. Investigating alternative CoT techniques, hybrid reasoning frameworks, and other optimization methods for both closed-source and open-source models represents an open challenge for future research. We hope that KRETA serves as a stepping stone for future advancements in this area, guiding the development of more effective reasoning strategies and robust VLMs.

\section*{Acknowledgements}

This work was supported by the Korea Institute for Advancement of Technology (KIAT) grant funded by the Ministry of Education, Korea Government, and by Seoul National University (Semiconductor-Specialized University).

\bibliography{custom}

\newpage
\appendix
\section{Appendix}

\begin{figure*}
  \centering 
  \includegraphics[width=0.85\textwidth]{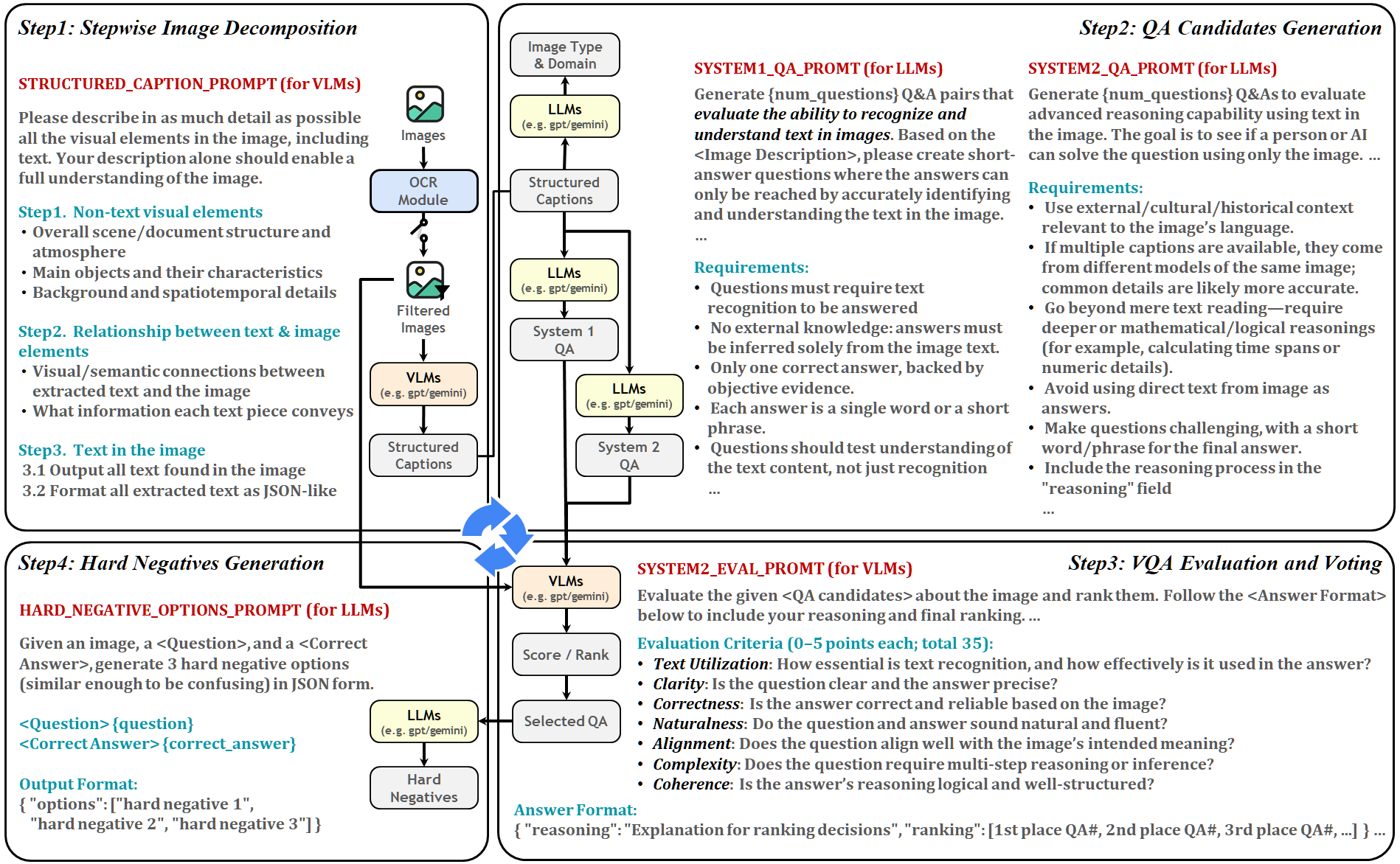} 
  \caption{An overview of the semi-automated VQA generation pipeline with prompts. Each step involves data processing using either VLMs or LLMs, with corresponding prompts shown in the figure. The actual data generation process uses Korean prompts. Prompts are shortened or omitted for readability.}
  \label{fig:auto-vqa-gen-pipeline} 
\end{figure*}

\subsection{Prompt for System 1/2 QA generation}

We release the full benchmark, including prompt files, dataset generation code, and evaluation scripts, at \href{https://github.com/tabtoyou/KRETA}{our official GitHub repository}.
In addition, the datasets and a public leaderboard with foundation model evaluation results are available on the \href{https://huggingface.co/datasets/tabtoyou/KRETA}{Hugging Face Hub}.

A compact overview of where each template is used in the pipeline appears in Figure~\ref{fig:auto-vqa-gen-pipeline}. For readability, we omit the full prompt text here; the exact templates and configurations are provided in the repository.

\subsection{Impact of Prompt Language on VQA Generation}

We observed that directly translating prompts into English \emph{while still requesting Korean outputs} consistently degraded QA quality compared to using native Korean prompts. We hypothesize three causes. 
First, there is a language-switching confusion effect whereby English instructions paired with Korean targets engage competing decoding pathways, reducing fidelity and format stability \cite{kim2025codeswitch}.
Second, literal translation often omits critical constraint cues and language-specific features such as discourse markers and honorific forms~\cite{enomoto-etal-2025-fair}.
Finally, fragmented Korean knowledge in current models is engaged most effectively when instructions, context, and outputs are all in Korean; English directives tend to trigger instruction-following patterns learned in English during fine-tuning, yielding suboptimal Korean outputs~\cite{acl23-xp3}.

Extending the pipeline to new languages should not rely on English templates. Instead, prompts should be native to the target language and tailored to its morphology, discourse markers, typographic conventions, and register to preserve both output quality and format stability.

\subsection{Image Types}
\label{app:image-types}
We organize 26 image types considered in KRETA. Table~\ref{tab:image_types} provides an overview of these types with concise descriptions.

\begin{table*}[t]
  \centering
  \resizebox{0.95\textwidth}{!}{%
  \begin{tabular}{l|p{12.5cm}}
    \toprule
    \textbf{Type} & \textbf{Description} \\
    \midrule
    \textit{Test paper} & Cropped exam items or full sheets containing questions, options, and instructions \\
    \textit{Mobile screenshot} & Smartphone app UI captures (buttons, tabs, banners, dialogs) \\
    \textit{Infographic} & Information graphics combining icons/figures with explanatory text \\
    \textit{Signage} & Indoor/on-site guidance signs (buildings, campuses, venues) \\
    \textit{Poster} & Event or promotional posters with title, schedule, and venue \\
    \textit{Menu} & Restaurant/café menus with item names, descriptions, and prices \\
    \textit{Book cover} & Front covers with title, author, publisher, and design elements \\
    \textit{PC screenshot} & Desktop/web application UI captures on a computer screen \\
    \textit{Store sign} & Storefront signage emphasizing brand/store names \\
    \textit{Chart and plot} & Bar/pie/line plots with axes, legends, and labels \\
    \textit{Public signs} & Official notices, warnings, and pictogram-based signs \\
    \textit{Banner} & Long horizontal/vertical banners for events or promotions, typically outdoor \\
    \textit{Street signs} & Traffic and directional signs in road contexts \\
    \textit{Product detail} & Product sheets/cards with specs, options, or feature callouts \\
    \textit{Table} & Rectangular grid layouts where key content resides in cells \\
    \textit{Manual} & User guides or how-to documents with procedural text and diagrams \\
    \textit{Presentation slides} & Slides (e.g., PPT/Google Slides) with titles, bullets, and figures \\
    \textit{Brochure} & Pamphlets/leaflets with sections, images, and promotional text \\
    \textit{Diagram} & Schematics or process/flow diagrams with labeled parts and arrows \\
    \textit{Document} & Structured textual documents not captured by other types \\
    \textit{Packaging label} & Labels on packaged goods with product name, ingredients, expiry, etc. \\
    \textit{Video scene} & Frames with on-screen text or subtitles (TV, film, online video) \\
    \textit{Receipts and invoices} & Printed proofs of purchase with line items, totals, and metadata \\
    \textit{Book page} & Inner pages of books (headings, paragraphs, footnotes, page numbers) \\
    \textit{Tickets and boarding passes} & Admission/transportation tickets with seat/time identifiers \\
    \textit{Handwriting} & Handwritten notes, memos, chalkboard/whiteboard writings \\
    \bottomrule
  \end{tabular}
  }
  \caption{The 26 \emph{image types} used in \textbf{KRETA}. Each type is characterized by its visual structure and the way textual information is organized and consumed. The taxonomy covers a wide spectrum of real-world materials, ranging from exam papers and screenshots to signage, diagrams, packaging, and handwritten notes, thereby reflecting the diverse document formats encountered in practical scenarios.}
  \label{tab:image_types}
\end{table*}

\end{document}